\documentclass[runningheads]{llncs}
\usepackage[T1]{fontenc}
\usepackage{graphicx}
\usepackage{amsmath}
\usepackage{amssymb}
\usepackage{bm} 

\usepackage{bm}
\usepackage{fix-cm}
\DeclareMathAlphabet{\mathsfit}{T1}{\sfdefault}{\mddefault}{\updefault}
\SetMathAlphabet{\mathsfit}{bold}{T1}{\sfdefault}{\bfdefault}{\updefault}

\newcommand{\vect}[1]{{\bm{\mathit{#1}}}}
\newcommand{\matr}[1]{{\boldsymbol{\mathsf{#1}}}}

\usepackage{xcolor}
\usepackage{lipsum}

\begin{document}
\title{Online model learning with data-assimilated reservoir computers}
%
%
\author{
Andrea N{\'o}voa \inst{1} \orcidID{0000-0003-0597-8326} 
\and Luca Magri \inst{1-3} \orcidID{0000-0002-0657-2611}}
\authorrunning{A. Nóvoa \and L. Magri}
%
\institute{
Imperial College London, Aeronautics Dept., SW7 2AZ, London, UK.
\and
The Alan Turing Institute, NW1 2DB, London, UK.
\and
Politecnico di Torino, DIMEAS, Corso Duca degli Abruzzi, 24 10129 Torino, Italy. 
\\
\{\email{a.novoa,l.magri}\}\email{@imperial.ac.uk}}
\maketitle              

\begin{abstract}
    We propose an online learning framework for forecasting nonlinear spatio-temporal signals (fields).  
    The method integrates 
    (i) dimensionality reduction, here, a simple proper orthogonal decomposition (POD) projection;
    (ii) a generalized autoregressive model to forecast reduced dynamics, here, a reservoir computer;
    (iii) online adaptation to update the reservoir computer (the model), here, ensemble sequential data assimilation.   
    We demonstrate the framework on a wake past a cylinder governed by the Navier-Stokes equations, exploring the assimilation of full flow fields  (projected onto POD modes) and sparse sensors. Three scenarios are examined: 
    a naïve physical state estimation; 
    a two-fold estimation of physical and reservoir states; and 
    a three-fold estimation that also adjusts the model parameters. 
    The two-fold strategy significantly improves ensemble convergence and reduces reconstruction error compared to the naïve approach.  
    The three-fold approach enables robust online training of partially-trained reservoir computers, overcoming limitations of \textit{a priori} training.   
    By unifying data-driven reduced order modelling with Bayesian data assimilation, this work opens new opportunities for scalable online model learning for nonlinear time series forecasting.

\end{abstract}
\section{Introduction}

Echo State Networks (ESNs) are effective reduced-order modelling and nonlinear time series forecasting tools~\cite{pathak_model_2018}. As reservoir computing systems, ESNs are generalized auto-regressive models that capture temporal correlations in data~\cite{aggarwal2018neural}. 
%
In fluid dynamics, ESNs have been successfully employed to model turbulent systems, capturing  their dynamical and predicting extreme events
 ~\cite{Racca_Doan_Magri_2023}. 
However, ESNs struggle with high-dimensional spatio-temporal flows and tend to lose accuracy over long prediction horizons, thus hindering their real-time modelling capabilities~\cite{pathak_model_2018,goswami2021data}.  
%
First, to address the dimensionality limitation, ESNs have been integrated with reduced order modelling techniques. These encode high-dimensional datasets into a reduced representation, either linearly, such as proper orthogonal decomposition (POD)~\cite{sharifi_echo_2024}, nonlinearly such as  convolutional autoencoders~\cite{doan_auto_2021,Racca_Doan_Magri_2023,ozalp_stability_2025}, then, the ESN predicts the evolution of the reduced system.  
%
Second, to improve the real-time predictability, ESNs  have been coupled with sequential data assimilation (e.g., the ensemble Kalman filter) in a Bayesian framework to update their predictions with sparse and noisy data that become available in real time~\cite{evensen_data_2009}. 
%
Current methods improve short-term forecasts, but do not update the ESN's internal states or parameters, limiting their generalizability~\cite{goswami2021data}. 
To overcome this limitation, we propose a method that extends the use of ensemble filters beyond state estimation in reservoir computers. We exploit the augmented state-space formulation~\cite{novoa_magri_2022} to seamlessly update the ESN's state predictions and trainable parameters in real time.   

%
%
%
%

\section{Methodology}

The test case is a two-dimensional unsteady laminar wake behind a cylinder at Reynolds number 100~\cite{Mo_Traverso_Magri_2024} (see Fig.\ref{fig:reconstruct}a). We refer to the noise-free dataset as the \textit{truth}, which is reserved for post-processing purposes only.
We add  Gaussian noise $\eta(\vect{x})=(\epsilon \ast \vect{G})(\vect{x})$ to the truth, where  $\epsilon\sim\mathcal{N}(0, 0.1^2)$ is convoluted with the Gaussian kernel $\vect{G}$ with standard deviation of 0.1.  
The noisy data (Fig.\ref{fig:reconstruct}b) are used to train the POD-ESN, and for online model updates via  data assimilation to simulate realistic conditions.

\subsection{Non-intrusive reduced-order model: POD-ESN}

\begin{figure}[htbp]
    \centering
    \includegraphics[width=\textwidth]{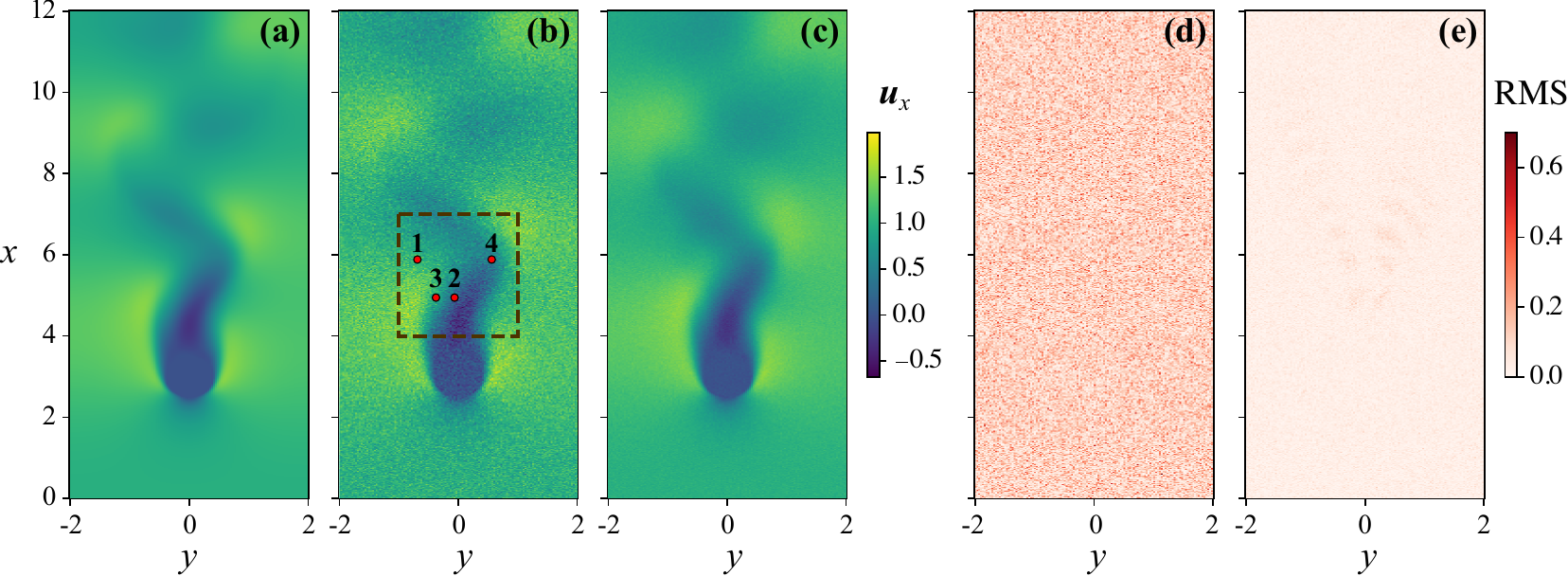}
    \caption{Streamwise velocity field of cylinder flow. Snapshots of (a) the truth; (b) the noisy data, which are the observable fields; (c) the POD reconstruction with $N_{m}=4$; and the instantaneous root mean squared (RMS) error field of panel (c) with respect to  (d) the noisy data, and (e) the truth. 
    The circles in (b) indicate the sensors, which are selected and numbered by the QR decomposition.}
    \label{fig:reconstruct}
\end{figure}

We propose a POD-ESN model as a non-intrusive reduced order model.   
First, we apply snapshot POD\footnote[1]{In this work, we use POD to keep the model as simple as possible; however, the proposed method can be extended to nonlinear model order reduction techniques.} to our dataset $\matr{U}=[\vect{u}_{x, 0}, \dots,  \vect{u}_{x, N_t}; \vect{u}_{y, 0}, \dots,  \vect{u}_{y, N_t}]$, where $N_t$ is the number of training snapshots, and $\vect{u}_{x}, \vect{u}_{y}$ are the velocity fields in the spatial coordinates $\vect{x}=[x;y]$. The snapshot POD decomposition is~\cite{poletti_modulo_2024} 
\begin{equation}
    \matr{U}(\vect{x}, t) = \matr{\Phi}(t)\matr{\Sigma}\matr{\Psi}^\mathrm{T}(\vect{x}),
\end{equation}
where 
$\matr{\Phi}$ are the temporal coefficients, 
$\matr{\Sigma}$ is the matrix of singular values, 
and $\matr{\Psi}$ is the orthonormal basis of POD modes. 
To reduce the dimensionality of the system we retain the first $N_{m}=4$ temporal modes,  $\vect{\phi}=[\phi_1;\dots;\phi_{N_{m}}]$, which contain 98\% of the energy of the true field~\cite{Mo_Traverso_Magri_2024}.  
Figure~\ref{fig:reconstruct} shows the reduced representation of the noisy field, which accurately reproduces the unseen true field.  
Second, we train an ESN to learn the time evolution of the temporal coefficients $\vect{\phi}$.  
The architecture is shown in Fig.~\ref{fig:schematic}. 
There are three key matrices in an ESN: (i) the input matrix $\matr{W}_{\mathrm{in}}$, which is scaled by $\sigma_\mathrm{in}$, maps the state, normalized by the factor $\vect{g}$, into the high-dimensional reservoir state $\vect{r}$;  (ii) the reservoir matrix $\matr{W}$, scaled by the spectral radius $\rho$, acts as a memory; and (iii) the output matrix $\matr{W}_\mathrm{out}$, which maps the reservoir state $\vect{r}$ into the physical domain.  
With this, the ESN equations used to forecast the POD coefficients are 
\begin{align}\nonumber
\label{eq:ESN_OG}
        \vect{\phi}_{k+1} &= \matr{W}_{\mathrm{out}}\left[\vect{r}_{k+1}; 1\right]\\
        \vect{r}_{k+1} &= \textrm{tanh}\left(\sigma_\mathrm{in}\matr{W}_{\mathrm{in}}\left[{\vect{\phi}}_k\odot\vect{g}; \delta\right]+
        \rho\matr{W}\vect{r}_k\right),
\end{align}
where the ${\tanh(\cdot)}$ is performed element-wise, $\odot$ is the Hadamard product, and $\delta=0.1$ is a symmetry-breaking constant~\cite{huhn_learning_2020}.  
Training the ESN consists of finding the weights of the output matrix $\matr{W}_\mathrm{out}$, while  $\matr{W}_{\mathrm{in}}$ and  $\matr{W}$ are sparse and randomly generated matrices that are fixed~\cite{lukovsevivcius_practical_2012}. The training consists of solving the linear regression problem
\begin{equation}
\label{eq:RidgeReg_ens}
    (\matr{R}\matr{R}^\mathrm{T} + \lambda_T \mathbb{I})\matr{W}_\mathrm{out}^\mathrm{T} = \matr{R}\matr{\Phi}^\mathrm{T}_\mathrm{train}, 
\end{equation}
where 
$\matr{\Phi}_\mathrm{train}$ is the  training dataset, 
$\matr{R}$ are the reservoir states corresponding to the training set, 
$\mathbb{I}$ is the identity matrix, 
and $\lambda_T$ is the Tikhonov regularization factor. 
The hyperparameters $\rho$ and $\sigma_\mathrm{in}$ are optimized with a recycle validation and Bayesian optimization. 
(For further details on the ESN training see~\cite{racca_robust_2021,novoa_magri_real_DT_2024}.)

\subsection{Online learning with data assimilation}

We wish to update the  POD-ESN model any time that data from sensors, which may be noisy and sparse, become available. This calls for real-time data assimilation. 
We focus on ensemble Kalman filters because they (i) can handle nonlinearities, and (ii) assimilate data on the fly. This allows us to achieve real-time modelling using the nonlinear ESN equations.  
\begin{figure}[!htb]
    \centering\includegraphics[width=.85\linewidth]{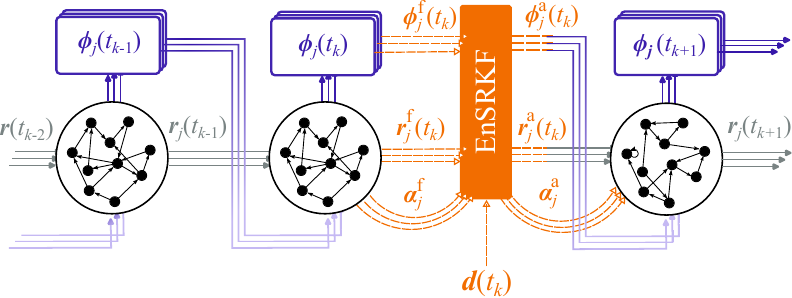}
    \caption{Schematic of online learning with echo state networks. The ESN model forecasts autonomously the POD coefficients $\vect{\phi}_k$ and the reservoir state $\vect{r}_k$. At the measurement time $t_k$, the EnSRKF updates $\vect{\phi}^\mathrm{f}_k$ and, optionally, $\vect{r}^\mathrm{f}_k$ and  time-constant parameters $\vect{\alpha}$. }
    \label{fig:schematic}
\end{figure}
From a statistical perspective, 
real-time data assimilation finds the most likely state ${\vect{z}}^\mathrm{a}$ of the system that is consistent with both the observed  data $\vect{d}$ and the model prediction ${\vect{z}}^\mathrm{f}$. We use the superscripts `a' and `f' to denote analysis and forecast, respectively, following the data assimilation nomenclature.  
In the ensemble approach, ${\vect{z}}^\mathrm{f}$ is estimated by propagating an ensemble of states of size $m$, and computing the ensemble statistics $\mathbb{E}({\vect{z}})\approx\bar{{\vect{z}}}=\frac{1}{m}\sum_j\hat{\vect{z}}_j$, and $\matr{C}_{zz}=\frac{1}{m-1}\sum_j({\vect{z}}_j-\bar{\vect{z}})({\vect{z}}_j-\bar{{\vect{z}}})^\mathrm{T}$.  
We define the augmented state  $\hat{\vect{z}}=[\vect{z}; \mathcal{M}(\vect{z})]$, where $\mathcal{M}:\mathbb{R}^{N_z}\rightarrow\mathbb{R}^{N_d}$ is the measurement operator mapping the state space to the observation space of size $N_d$. 
The observations $\vect{d}$ are measurements from the $\vect{u}_x$ and $\vect{u}_y$ fields. We compare the assimilation of the full flow field projected on the POD modes, i.e., the POD coefficients and $\mathcal{M}(\vect{z}) \triangleq \vect{\phi}$,  with the assimilation of sparse measurements  such that $\mathcal{M}(\vect{z}) \triangleq [\vect{u}_x({\vect{x}}_q); \vect{u}_y({\vect{x}}_q)]$, where $\vect{x}_q$ is the location of the sensors (Fig.\ref{fig:reconstruct}b). 
When observations become available, the ensemble is updated as~\cite{evensen_data_2009}
\begin{align}\label{eq:EnKF}
    \hat{\vect{z}}_j^\mathrm{a} 
    &= \hat{\vect{z}}_j^\mathrm{f}+\matr{C}_{\hat{z}\hat{z}}^\mathrm{f}\matr{M}^\mathrm{T}\left(\matr{C}_{dd}+\matr{M}\matr{C}_{\hat{z}\hat{z}}^\mathrm{f}\matr{M}^\mathrm{T}\right)^{-1}\left(\vect{d}_j - \matr{M}\hat{\vect{z}}_j^\mathrm{f}\right),
\end{align}
where $\matr{M}=[\matr{0}\mid\mathbb{I}]$, and $\matr{C}_{dd}$ is the prescribed measurement error covariance. 
We solve equations~\eqref{eq:EnKF} via the square-root transform, i.e., with the ensemble square root  Kalman Filter (EnSRKF). 
The EnSRKF can handle non-Gaussian error statistics and nonlinear dynamics and often requires smaller ensembles than traditional Kalman filters~\cite{tippett_ensemble_2003}. 
The online learning process is illustrated in Figure~\ref{fig:schematic}. We compare three approaches:
\begin{enumerate}
    \item \textit{Physical state estimation}: update the ESN prediction of the POD coefficients, ${\vect{z}} \triangleq \vect{\phi}$. This naïve estimation is equivalent to performing an open-loop step.  
    \item \textit{Two-fold state estimation}: update simultaneously the POD coefficients and the reservoir state (the full state) by augmenting the state vector ${\vect{z}} \triangleq [\vect{\phi}; \vect{r}]$.  
    \item \textit{Three-fold state and parameter estimation}: update simultaneously the full state and the trainable parameters of the ESN, i.e., the output matrix.   
    We factorize  $\matr{W}_\mathrm{out}$  using singular value decomposition to  
    form  $\matr{W}_\mathrm{out} = \matr{X}\matr{A}\matr{V}^\mathrm{T}$, where $\matr{A}$ is the diagonal matrix of singular values of $\matr{W}_\mathrm{out}$.  
    With this, rather than modifying every  entry in $\matr{W}_\mathrm{out}\in\mathbb{R}^{N_r\times N_{m}}$, we update the $N_{m}$  singular values $\vect{\alpha}$, and ${\vect{z}} \triangleq[\vect{\phi}; \vect{r}; \vect{\alpha}]$.
\end{enumerate} 

\section{Results}

\begin{figure}
    \centering
    \includegraphics[width=\textwidth]{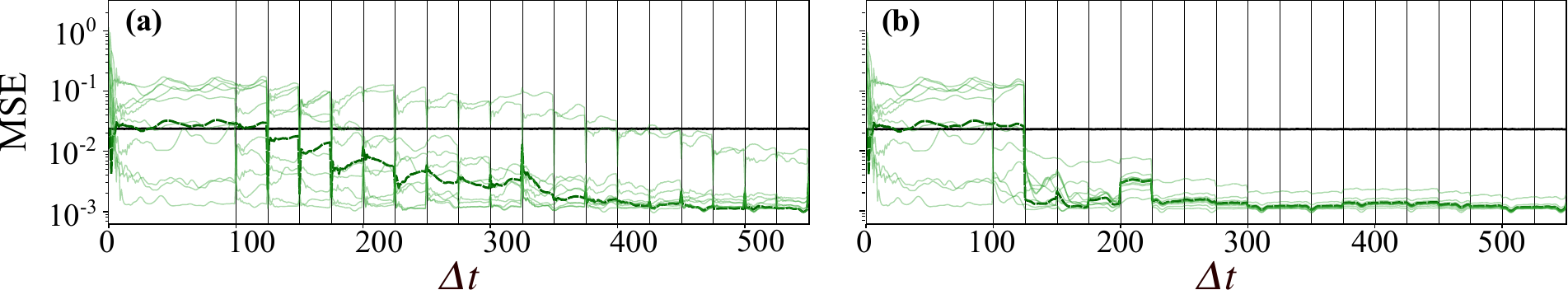}
    \caption{Comparison between (a) physical state estimation, and (b) two-fold state estimation. Time evolution of the mean squared error (MSE) for each ensemble member (thin lines) and their mean (dashed line), and the noisy data (black). }
    \label{fig:mse_state}
\end{figure}

\begin{figure}
    \centering
    \includegraphics[width=\textwidth]{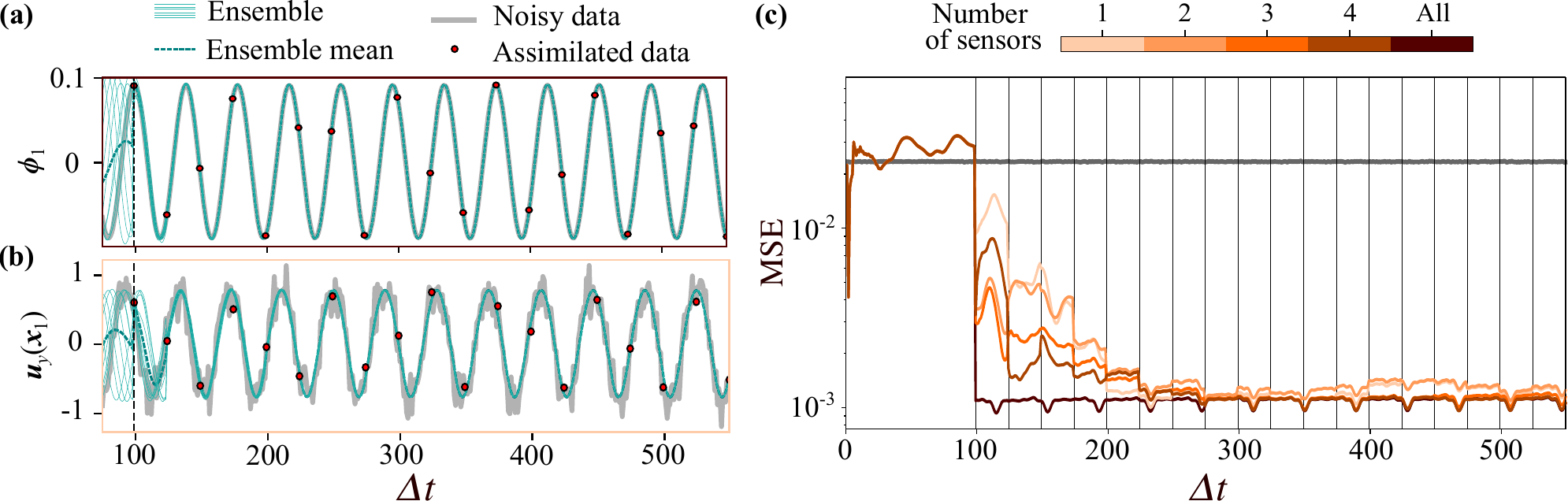}
    \caption{Effect of the number of sensors on the full state estimation. Comparison between the evolution of the ensemble (blue) forecast and the reference data (gray) of (a) the first POD coefficient for assimilation of the full fields, and (b) the span-wise velocity for $N_d = 1$. (c) MSE of the ensembles (colormap), and the noisy data (black). The assimilation times are indicated with red dots (a-b) and vertical lines (c).}
    \label{fig:Nq}
\end{figure}

We train a POD-ESN with $N_r=40$ on  $N_t=250$ snapshots, which corresponds to six periods of the principal POD coefficient $\phi_1$. The ESN hyperparameters $\rho= 0.976$ and $\sigma_\mathrm{in} = 0.890$, are optimized in $[0.5,50]\times[0.2,1.05]$ for $[\sigma_\mathrm{in}\times\rho]$ (in logarithmic scale for $\sigma_\mathrm{in}$). 
%
First, we compare in Figure~\ref{fig:mse_state} the physical state estimation and the two-fold estimation of both physical and reservoir states with an ensemble with $m=10$, and observations from a single sensor (number~1 in Fig.~\ref{fig:reconstruct}) every $\Delta=\Delta t_{d} / \Delta t=25$. 
We analyse the mean squared error (MSE) of the POD-ESN prediction $\vect{u}^\star$   with respect to the true velocity fields
\begin{equation}
    \mathrm{MSE} = \dfrac{1}{2N_xN_y} \left(||\vect{u}^\star_x-\vect{u}_x^\mathrm{truth}||^2 +  ||\vect{u}^\star_y-\vect{u}_y^\mathrm{truth}||^2 \right)
\end{equation}
where $||\cdot||$ denotes the $L_2$-norm and $N_x N_y=66177$ are the number of grid points in the domain.  
The ensemble converges faster with smaller uncertainty (represented by the ensemble spread) for full state assimilation (Fig.~\ref{fig:mse_state}). This is because not only we perform an open loop step, but we also update the reservoir state.  Hence, we focus on the full state for the remainder of this work. 
Figure~\ref{fig:Nq} shows the effect of the number of observations  $N_d$ in the reconstruction.  With full observability, the error decreases suddenly  with the first observation because we have direct information in the reduced dimension (Fig.~\ref{fig:Nq}a), and the projection onto the POD modes decreases the noise from the field measurement (Fig.~\ref{fig:reconstruct}). 
With partial observations, the more sensors, the faster the convergence, and, even when $N_d=1$, the MSE is as small as it would be with full observability. 
This means that our trained POD-ESN is a \textit{good} model of the system. 

\begin{figure}[!hb]
    \centering
    \includegraphics[width=\textwidth]{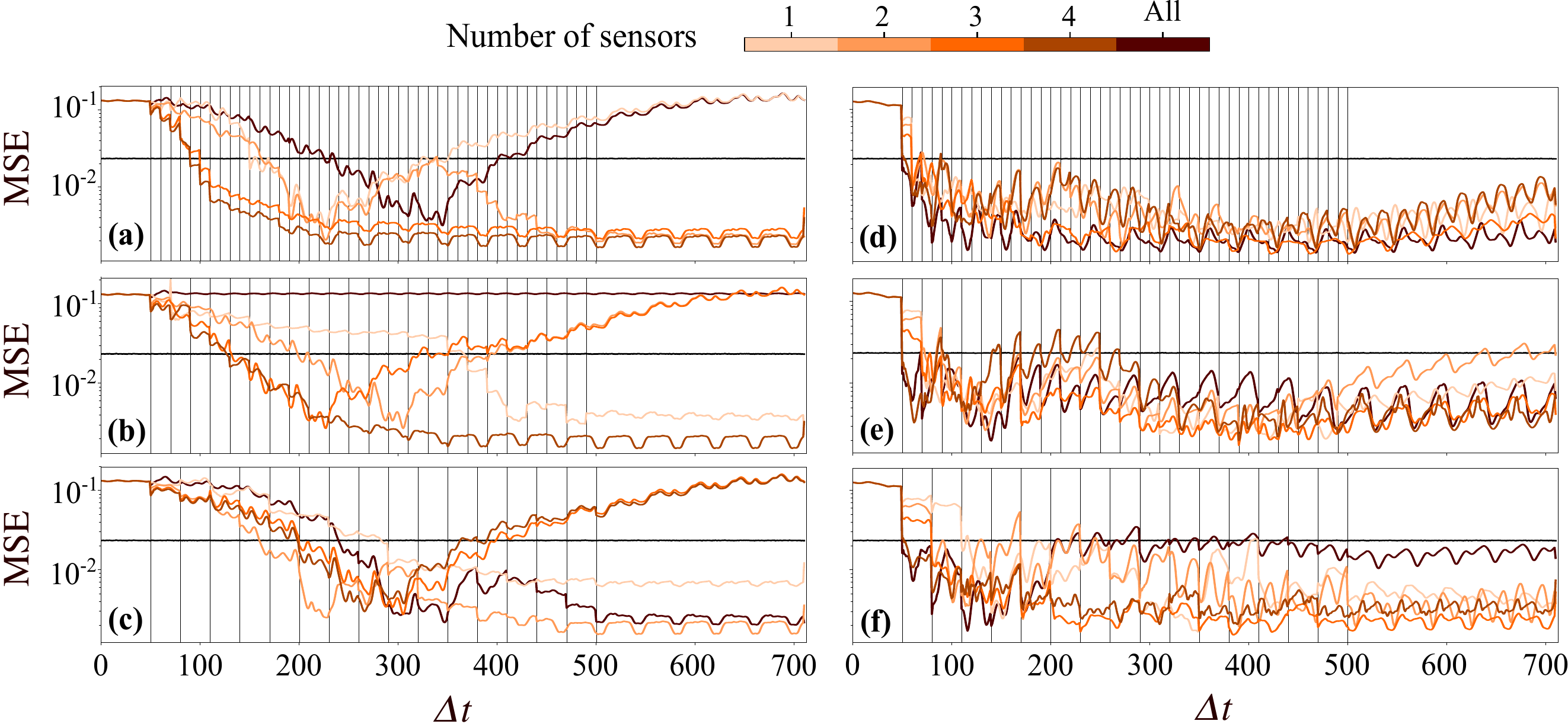}
    \caption{Comparison between (a-c) two-fold and (d-f) three-fold  estimation for varying number of sensors. Reconstruction MSE of the ensemble mean (colormap), and the noisy data (black). 
    The vertical lines indicate the assimilation steps such that (a,d)  $\Delta = 10$, (b,e)  $\Delta = 20$, and (c,f)  $\Delta = 30$. 
    }
    \label{fig:MSE}
\end{figure}

Next, we train a POD-ESN with parameters identical to earlier experiments but on a truncated training set of  $N_t=90$ (2.1 periods of  $\phi_1$). This yields a partially-trained POD-ESN, which does not fully capture the statistics of the system. To address this, we perform online parameter adaptation with the EnSRKF, i.e., we continue training the ESN on the fly.   
%
The EnSRKF converges to a consistently similar set of $\vect{\alpha}$ with $m>10$ (results not shown for brevity).  
A larger ensemble is needed due to the increased uncertainty and degrees of freedom. 
In Figure~\ref{fig:MSE}, we set $m=50$ and compare the online learning with (Fig.~\ref{fig:MSE}~a-c) and without (Fig.~\ref{fig:MSE}~d-f) parameter estimation for varying $\Delta$ and $N_d$. 
On the one hand, the two-fold estimation often yields MSEs higher than the noisy data MSE. This is due to an early collapse of the ensemble covariance, i.e.,  the model is overfitting to the initial observations. 
On the other hand, the three-fold estimation 
avoids the covariance collapse by dynamically adapting the singular values of the ESN output matrix, thereby maintaining ensemble diversity and uncertainty quantification.
This demonstrates that joint state-parameter estimation enables the POD-ESN to generalize beyond the training horizon, with the EnSRKF effectively updating $\matr{W}_\mathrm{out}$ to learn the system dynamics. 
%

\section{Conclusions}

We introduce an online learning approach for forecasting spatio-temporal fields that integrates real-time data assimilation with data-driven  modelling. The proposed strategy combines: 
(i) a dimensionality reduction technique, 
(ii) a generalized autoregressive model, and 
(iii) an online data assimilation tool.  

Specifically, we use snapshot proper orthogonal decomposition (POD) to reduce the dimensionality of the system; we train a reservoir computer---an echo state network (ESN)---to model the temporal evolution of the POD coefficients; and  we update the POD-ESN on the fly via the ensemble square-root Kalman filter (EnSRKF). 
We demonstrate the method on a two-dimensional unsteady laminar wake behind a cylinder governed by the Navier-Stokes equations, using noisy velocity flow field measurements for both training and online learning. 
Two data assimilation strategies are explored: one that assimilates the full field projected onto the POD modes, and another that uses sparse sensors in the wake. 
For each strategy, we test three incremental online learning scenarios for the ESN: 
(i) a naïve estimation of the physical state (the POD coefficients), 
(ii) a two-fold estimation of the POD coefficients and the reservoir states (the memory of the network), and
(iii) a three-fold estimation of the POD coefficients, reservoir states, and model parameters---specifically, by updating the singular values of the ESN's trainable output matrix, $\matr{W}_\mathrm{out}$. 

We find that simultaneously updating both the physical and reservoir states reduces reconstruction error and enhances filter convergence, even with limited observability, which validates the robustness of the trained POD-ESN. Furthermore, the EnSRKF can optimally update $\matr{W}_\mathrm{out}$ in a Bayesian framework alongside the physical and reservoir states, therefore enabling online training of the ESN, which is of significant advantage when handling partially trained data-driven models.
This work bridges the gap between offline training and real-time forecasting, opening new opportunities for online learning in nonlinear time series forecasting.

\subsubsection{\ackname} We acknowledge the support from the UKRI AI for Net Zero grant EP/Y005619/1, the ERC Starting Grant PhyCo 949388, and the grant EU-PNRR YoungResearcher TWIN ERC-PI\_0000005.

\bibliographystyle{splncs04}
\bibliography{mybibliography}
\end{document}